\documentclass{article}

\usepackage{microtype}
\usepackage{graphicx}
\usepackage{subfigure}
\usepackage{booktabs} 

\usepackage{hyperref}
\usepackage{multirow} 
\usepackage{makecell}

\usepackage[accepted]{icml2025}

\usepackage{amsmath}
\usepackage{amssymb}
\usepackage{mathtools}
\usepackage{amsthm}
\usepackage{graphicx}
\usepackage[capitalize,noabbrev]{cleveref}

\theoremstyle{plain}

\theoremstyle{definition}

\theoremstyle{remark}

\usepackage[textsize=tiny]{todonotes}

\icmltitlerunning{Submission and Formatting Instructions for ICML 2025}

\begin{document}

\twocolumn[
\icmltitle{Revisiting Large Language Model Pruning using Neuron Semantic Attribution}



\icmlsetsymbol{equal}{*}

\begin{icmlauthorlist}
\icmlauthor{Yizhuo Ding}{yyy}
\icmlauthor{Xinwei Sun}{yyy}
\icmlauthor{Yanwei Fu}{yyy}
\icmlauthor{Guosheng Hu}{comp}
\end{icmlauthorlist}

\icmlaffiliation{yyy}{Fudan University}
\icmlaffiliation{comp}{University of Bristol}

\icmlkeywords{Machine Learning, ICML}

\vskip 0.3in
]


\printAffiliationsAndNotice{}  

\begin{abstract}

Model pruning technique is vital for accelerating large language models by reducing their size and computational requirements. However, the generalizability of existing pruning methods across diverse datasets and tasks remains unclear. Thus, we conduct extensive evaluations on 24 datasets and 4 tasks using popular pruning methods. Based on these evaluations, we find and then investigate that calibration set greatly affect the performance of pruning methods. In addition, we surprisingly find a significant performance drop of existing pruning methods in sentiment classification tasks. To understand the link between performance drop and pruned neurons, we propose Neuron Semantic Attribution, which learns to associate each neuron with specific semantics. This method first makes the unpruned neurons of LLMs explainable. 

\end{abstract}
\section{Introduction}
Large Language Models (LLMs) have made significant strides in a wide range of Natural Language Processing (NLP) tasks, including text generation, sentiment classification, machine translation, and question answering. The success of LLMs, such as GPT-3 and 4 \cite{achiam2023gpt}, BERT \cite{devlin2018bert}, and LLaMA families \cite{touvron2023llama}, has been a key milestone in the development of AI. These models have demonstrated a remarkable ability to understand and generate human-like text, significantly improving applications in various industries, including healthcare, finance, customer support, and content creation. The advancements made by LLMs are not only transformative for the AI community but also promising in improving productivity, enabling new technologies, and improving decision making in many sectors.

However, despite their extraordinary capabilities, LLMs come with a clear  drawback—high computational cost during inference. Deploying these models for inference is resource-intensive, requiring substantial memory and processing power. This results in high energy consumption, making it difficult for those with limited resources to fully leverage these models for practical applications.

Given the high costs associated with LLMs, reducing the computational burden while preserving their performance is critical. 

To address this challenge, model pruning \cite{frantar2023sparsegpt,sun2023simple,zhang2024plug} is an effective way of accelerating LLMs. It involves removing unnecessary weights or neurons from the model, thereby reducing its size and computational cost. Based on the granularity of the components being removed, pruning methods can be categorized as structured and unstructured pruning. Structured pruning removes entire components of the model, such as attention heads, channels, or even layers, resulting in a smaller and more efficient architecture. This approach is easier to implement on hardware since it reduces the overall computational graph. Unstructured pruning, on the other hand, operates at the individual weight level, selectively removing parameters that contribute the least to the model’s performance. Most pruning methods for LLMs are conducted without training, i.e. post-training pruning, due to the simplicity and effectiveness. This method does not require fine-tuning the model after pruning, meaning that no additional updates to the model parameters are necessary. As a result, it is computationally inexpensive and fast.

Despite the great success of pruning methods \cite{frantar2023sparsegpt,sun2023simple,zhang2024plug}, very little research extensively investigates the generalizability of the pruned models on extensive tasks and datasets. However, we believe the generalizability is very important for pruned models. The work \cite{bandari2024c4} evaluated 2 pruned models on 9 datasets and 3 tasks. In addition, another work \cite{williams2024impact} evaluated 9 pruned models on 10 datasets and 4 tasks. However, more tasks and data sets are expected to be evaluated to test the generalizability of pruned models. To bridge this gap, we conduct a more extensive evaluations on 10 models (Llama-2-7b, Llama-2-13b, Llama-3-8b \cite{touvron2023llama}, OPT-6.7b, OPT-13b and OPT-30b \cite{zhang2022opt} etc), 24 datasets (SST2, WNLI, Yelp and ARC-C etc), and 4 major tasks (sentiment classification, question answering, semantic similarity and logical reasoning).

We also extensively evaluate the components of pruning methods including calibration data, sequence length of tokens and target sparsity ratio of prune methods. We find that the calibration data plays a significant role in performance. Though existing methods \cite{bandari2024c4} find the performance drop e.g. accuracy of Llama 2-Chat 7B model pruned with Wanda, drop from 53.78\% to 48.12\% on e-SNLI dataset for natural language inference with different calibration sets. However, the discovery and analysis of the impact of the calibration set have not been investigated deeply. For example, our extensive evaluations find the accuracy of OPT 7B model pruned with Wanda drop from 77.42\% to 55.08\% on Yelp dataset for sentiment classification.

We also unexpectedly find that certain tasks, such as sentiment classification, experience a significant performance drop when using existing pruning methods \cite{frantar2023sparsegpt,sun2023simple,zhang2024plug}. This intriguing result motivates us to investigate the underlying reasons behind this. During this investigation, we realize that understanding the connection between pruned neurons and performance drop proved to be a fundamental challenge. To bridge this gap, we propose Neuron Semantic Attribution (NSA) method which learns to assign semantics to each neuron, facilitating us to understand the meaning of the pruned (and also unpruned) neurons. Based on NSA, for example, we find the sentiment performance drop of compressed OPT-6.7b model because the pruned neurons exhibit associations with sentiment-linked semantics, such as `trust' and `honestly'.

\begin{figure*}[ht] 
    \centering
    \includegraphics[width=1.0\linewidth]{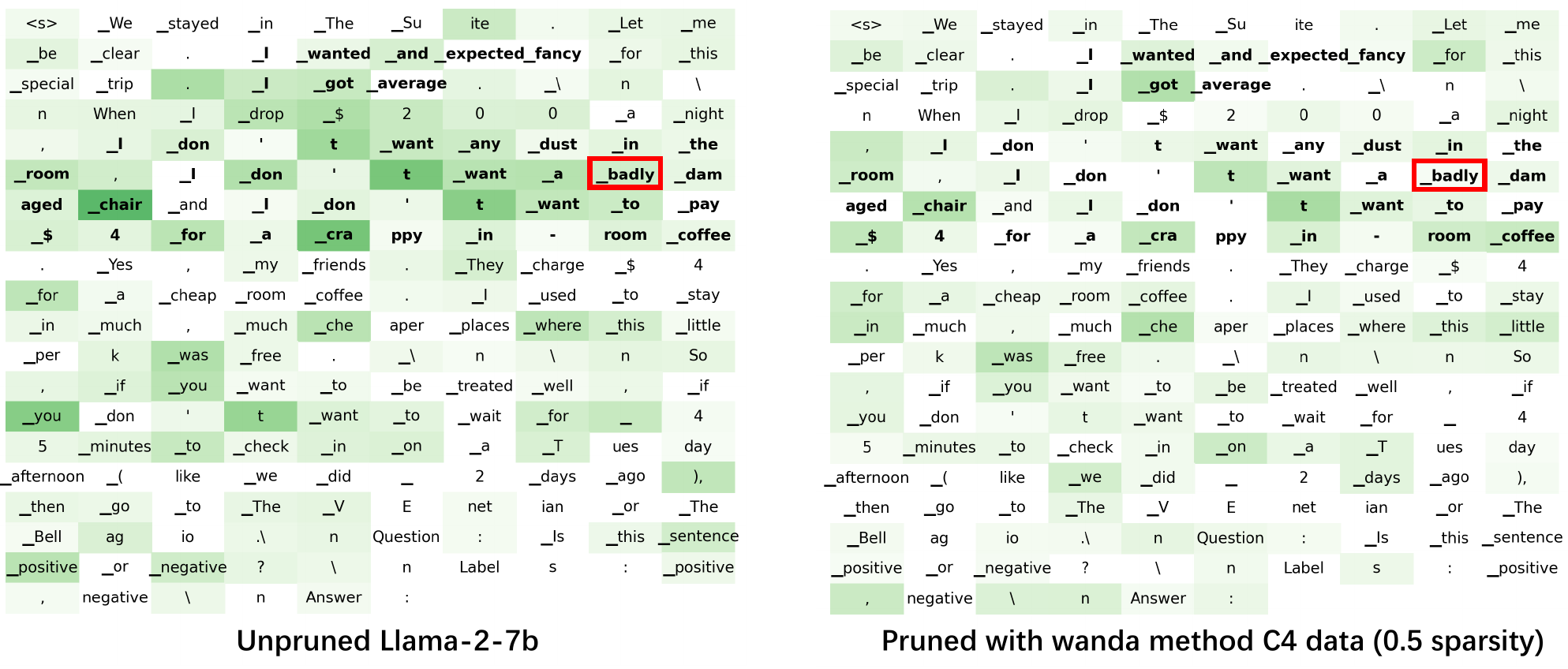}
    \vspace{-0.2in}
    \caption{Application of the NSA method on Yelp data, showing the activations of neuron 2194 in layer 15. Darker colors represent stronger  activations. For example, the activation of the pruned and unpruned models for the sentiment-related semantics 'badly' drops from 0.2112 to 0.0084, explaining the degraded sentiment classification of the pruned model.}
    \vspace{-0.1in}
\label{fig:Yelp_neuron}
\end{figure*}

This study makes several significant contributions to the field of model optimization for LLMs:
\begin{enumerate}
\item We conduct extensive evaluations of pruned models across a wide range of tasks, datasets, and models to better understand the generalizability of pruned models and bridging gaps in previous research. The systematic evaluations  have yielded valuable insights and provided crucial guidance for future research in this field.
    
\item We find that the calibration data plays a significant role in the performance of pruned models, particularly for tasks like sentiment classification. Our results highlight that calibration data should be tailored to specific tasks and datasets for optimal pruning performance.

\item To our knowledge, very little research investigates the explainability and semantics of the pruned neurons/components, which are actually important to understand and analyze the pruning methods. To bridge this gap, we propose a  Neuron Semantic Attribution method, which learns to assign specific semantics to  neurons. For example, in Figure \ref{fig:Yelp_neuron}, this method shows the pruned neurons are associated with semantics like `badly', providing a clear explanation for the drop of sentiment performance. 
\end{enumerate}

\section{Related work}

\textbf{Pruning Strategy} 
Model pruning has emerged as an effective and efficient method for compressing large neural networks, particularly in the context of LLMs. 

Recent advancements in  pruning methods for LLMs have introduced several promising methods. 
For instance, SparseGPT \cite{frantar2023sparsegpt} introduces a pruning method that incorporates OBS updates (optimal block sparsity), which allows for efficient pruning while retaining high accuracy in large models. This method revolutionized pruning by addressing both sparsity and computational efficiency in one framework. Following this, Wanda \cite{sun2023simple} introduces a simpler, yet highly effective pruning technique that further reduces the computational burden by refining the selection of parameters to prune, while still maintaining the model’s original functionality. RIA \cite{zhang2024plug} improves the pruning metric with relative importance and used channel permutation to adjust the pruned model. However, it is not clear the generaliaility of those methods for downstream tasks.

\textbf{Calibration Data} Calibration data refers to a small set of unlabeled samples used to adjust model parameters during model compression including model pruning and post-training quantization. In pruning process, calibration data helps guide the selection of parameters to prune, ensuring that critical model components are preserved. Studies \cite{frantar2023sparsegpt, sun2023simple, zhang2024plug} combine calibration data with model weight to calculate the pruning metrics. In quantization \cite{frantar2022optimal}, calibration data is utilized to set activation ranges, ensuring that the quantized model maintains performance close to the original. One work \cite{hubara2021accurate} demonstrated that even with a minimal calibration set, it is  possible to achieve accurate post-training quantization by optimizing layer parameters over this data. Recent studies show that for both pruning and quantization, proper calibration data selection helps maintain task performance, while poor selection may cause underperformance. 

\textbf{Generalizability}
It is interesting to explore the generalizability of pruned models. 
 Recent studies \cite{williams2024impact, bandari2024c4} have shown that calibration data can significantly impact the pruning process and pruned models' generalizability over various tasks. 
 These findings suggest that the pruning metrics derived from different calibration data may lead to varied outcomes, especially when applied to diverse downstream tasks like sentiment analysis and question answering. In particular, certain datasets may cause pruned models to underperform due to their inability to capture critical patterns that are necessary for specific tasks. Despite the existing study \cite{williams2024impact, bandari2024c4} on generalizability, 
  more extensive evaluations and analysis are needed. 
 Our work conduct extensive experiments with a variety of models, datasets, and tasks to explore the generalization capability of pruned models. Furthermore, we build upon these insights by introducing a novel visualization method that helps to better understand these impacts, which named Neuron Semantic Attribution. This method allows for in-depth analysis of neuron-level activations and their relationships with input tokens, offering new insights into the specific components of pruned models that contribute to model predictions.

\section{Methodology}

\begin{figure*}[!ht] 
    \centering
    \includegraphics[width=0.8\linewidth]{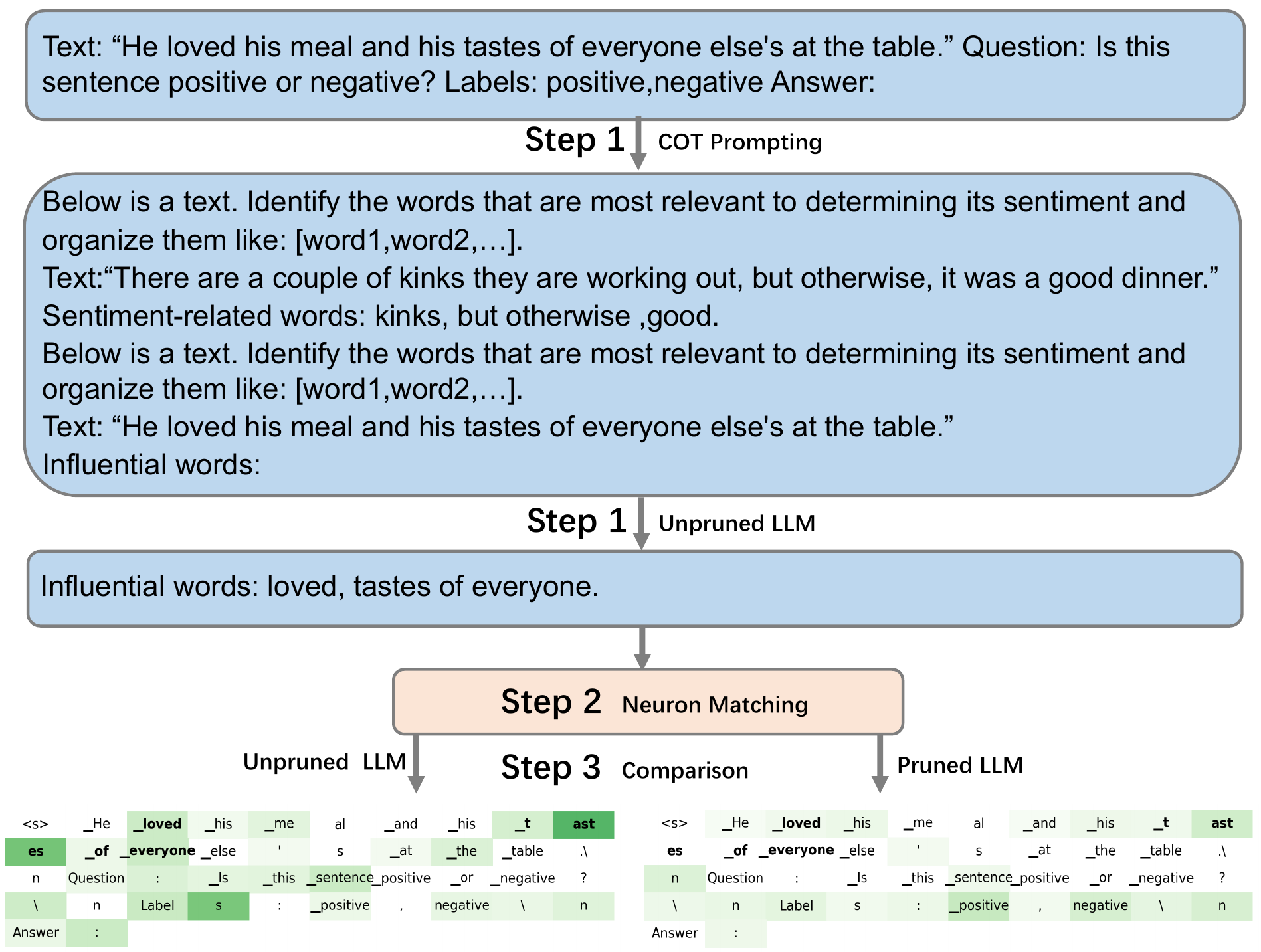}
    \vspace{-0.in}
    \caption{The framework of NSA. {Step 1, Influential Words  Selection; Step 2, Neuron Matching; Step 3,  Comparison.}}
    \vspace{-0.1in}
    \label{fig:framework}
\end{figure*}
\subsection{Pruning Methods}
\label{pruning_methods}

In this study, we employed three state-of-the-art post-training pruning methods: SparseGPT \cite{frantar2023sparsegpt}, Wanda \cite{sun2023simple}, and RIA \cite{zhang2024plug}. These methods are designed to reduce the size of large language models (LLMs) while maintaining their performance across various tasks. Even though all three methods prune the model using a combination of weights $W$ and activations $X$, they each use different metrics $M$ and strategies to decide which parameters to prune.

\begin{itemize}
    \item \textbf{SparseGPT}: This method selects parameters based on the metric: $M=\frac{W^2}{{[H^{-1}]}^2}$,where $H^{-1} = (XX^\top + \lambda I)^{-1}$. After pruning, SparseGPT also applies an OBS \cite{hassibi1993optimal} update to compensate for the pruned parameters.
    \item \textbf{Wanda}: This method uses a simple yet effective metric for pruning: $M=|W| \cdot \|X\|_2$. After pruning, there is no need for any parameter update step.
    \item \textbf{RIA}: This method calculates the pruning metric using relative importance: $M = \left( \frac{|W_{ij}|}{\sum |W_{*j}|} + \frac{|W_{ij}|}{\sum |W_{i*}|} \right) \times \left( \|X_i\|_2 \right)^a$. After calculating the metric, channel permutation is used to adjust the model for N:M sparsity.
\end{itemize}
These methods were evaluated on a range of tasks to assess how each performs across different calibration data. Each pruning method was applied iteratively, with models being pruned and evaluated to ensure that performance was maintained across tasks.

\subsection{Dataset, Task and Evaluation}
\label{model_pruning}

To investigate the effects of pruning methods on model generalization, we extensively evaluated three pruning methods.  Our experiments were conducted on a set of 14 large language models, including popular models such as OPT-6.7b, Llama-2-7b,DeepSeek-R1-qwen2-7b \cite{guo2024deepseek},Vicuna-7b \cite{chiang2023vicuna},Mistral-7b \cite{jiang2023mistral}, across 24 distinct datasets. Our evaluated datasets cover a diverse set of tasks and are grouped into four primary categories.  \textbf{Sentiment Classification}: SST2 \cite{socher2013recursive}, Yelp, IMDB \cite{rothe2018deep}, Sentiment140 \cite{go2009twitter}. The tasks were focused on classifying text based on the sentiment it expresses. \textbf{Question Answering}: ARC-C \cite{clark2018think}, ARC-E, OpenBookQA \cite{mihaylov2018can}, PubMedQA \cite{jin2019pubmedqa}, RACE \cite{lai2017race}, BoolQ \cite{clark2019boolq}, QNLI \cite{wang2018glue}. The tasks provided answers to specific questions based on provided context. \textbf{Semantic Similarity}: MRPC, QQP, WIC, SciTail \cite{khot2018scitail}, RTE, MedNLI 
 \cite{romanov2018lessons}, PAWS 
 \cite{yang2019paws}. The tasks assessed  the similarity between different pieces of text. \textbf{Logical Reasoning}: COPA 
 \cite{afshar2018copa}, LogiQA 
 \cite{liu2020logiqa}, SciQ 
 \cite{welbl2017crowdsourcing}, WNLI, WinoGrande 
 \cite{sakaguchi2021winogrande}, WSC. The tasks required  the model to reason logically and solve problems based on textual input.

The calibration data used in these experiments includes: C4, WikiText2, PTB, ARC-C \cite{clark2018think}, ARC-E \cite{clark2018think}, BoolQ, RTE, SST2, and WNLI. These data sets represented a diverse set of domains and tasks to ensure the robustness of the findings in various use cases. These various datasets were used to well explore the impact of different calibration data on pruning effectiveness.

The performance of each model was evaluated using the standard metrics, accuracy. Our evaluation process includes:(1) Model pruning: This involved setting different calibration data, sentence lengths, and sparsity ratios for the dense models. (2) Evaluation: After pruning, we tested the models on 24 different datasets.

By evaluating  these various tasks and datasets, we gain a comprehensive understanding of the  generalizability of the pruning methods.

\subsection{Neuron Semantic Attribution}
\label{nsa}
During our evaluations, we find it is fundamentally challenging to understand the unexpected performance drop of pruning methods on some tasks, e.g. sentiment classification. To our knowledge, the explainability of pruning methods in the aforementioned behaviors is greatly overlooked by the society. To bridge this gap, we propose a new method, Neuron Semantic Attribution (NSA). NSA can learn the relationship between neurons and tokens in the input text, allowing us to build the links between neurons and specific tokens. 

The NSA method include 3 steps. Step 1, we first sample some text (e.g. sentences) from datasets and smartly construct prompts using Chain-of-Thought (CoT) \cite{wei2022chain} to identify influential words for specific tasks (e.g. `unacceptable' and `angry' for sentiment classification). Step 2, we feed this text with influential words to the unpruned network to record the activations of neurons. The \emph{important} neurons corresponding to specific influential words are then found with high activations measured by a predefined score function. In this way, we can link the semantics (influential words) with specific neurons in this unpruned model. Step 3, the same text is then fed to the \emph{pruned} model, therefore, we can  compare the activations of the important neurons between the unpruned and pruned models. If the activations of one specific semantics (defined by influential words) significantly decrease,  we think the pruned model removes important neurons for specific tasks, potentially degrading the performance of that task. For example, the activation of `badly' in Figure \ref{fig:Yelp_neuron} significantly drops, potentially decreasing the performance of sentiment classification.

These steps are detailed as below:

\paragraph{Step 1: Influential Words Selection}
We begin by choosing a set of text examples from relevant datasets to analyze neuron activations as is shown in Figure \ref{fig:framework}. For each task, we select around 10 examples to ensure that they represent the task well. For example, we select data from the ARC-C dataset for question answering tasks and from the Yelp dataset for sentiment analysis tasks. Next, we apply the Chain-of-Thought (CoT) technique to identify influential words related to the task. The model reasons through the task and identifies words that significantly influence the decision-making process. As shown in Figure \ref{fig:Yelp_neuron}, for sentiment analysis tasks on the Yelp dataset, the model identifies words like `badly' and “damaged” as the most influential in determining sentiment.

\paragraph{Step 2: Neuron Matching} In this step, we match  neurons with specific semantics (influential words). We firstly  select neurons that have strong activations for all the influential words as a whole; secondly we attribute each selected neuron to specific influential words. Firstly, once the influential words are identified in Step 1, we feed the input text into the  unpruned  models. 
The activation values of each neuron corresponding to all tokens are then recorded. Given one neuron, after that, we calculate a normalized activation of influential words over all the words/tokens by:
\begin{align}
Score= \frac{\sum_{m \in S}A_{m}}{\sum_{n=1}^{N}A_{n}}   \label{Eq:nsa_step2}
\end{align}
where $A_m$ is the activation in   $S$ which is a set of tokens related to influential words. $N$ is the total number of tokens. Thus, we select the neurons with top $Score$ values, meaning that these neurons are crucial for a task with aforementioned influential words. Secondly, we match the neurons with specific influential words. 
Given one neuron selected, we can obtain its activation over the $m$-th influential word, i.e. $A_m$ from Eq. (\ref{Eq:nsa_step2}).
This given neuron can then be matched with the influential words with higher activations $A_m$. In this way, we successfully attribute the one  neuron with particular semantics (influential words with high $A_m$) as shown in Figure \ref{fig:Yelp_neuron}.

\paragraph{Step 3: Comparison of the Unpruned and Pruned}The same input text is then fed to the pruned model. We can calculate $A_m$ activations of the same selected neurons in Step 2 from the pruned model. After that, we can compare the activation maps of the unpruned and pruned models as shown in Figure \ref{fig:Yelp_neuron}. 
 A significant drop of $A_m$  suggests that the pruning method prunes some neurons which are highly related to the $m$th influential word, possibly leading to performance degradation.

\paragraph{Discussions} NSA is  compatible with various LLMs, including LLaMA and OPT, and provides users with the flexibility to select the data samples they wish to investigate. It also facilitates future research on the effects of pruning on large language models, offering a robust platform for understanding the impact of pruning on model performance. In addition, we continuously monitor the pruning process to ensure that the pruned models retain their performance across different tasks. By visualizing neuron activations, we can make informed decisions about the optimal pruning strategy and identify potential areas for improvement in future pruning approaches.

Moreover, NSA allows us to monitor task-specific performance more objectively. By examining the relationship between neuron activations and model performance, we can identify if pruning leads to unwanted biases or performance degradation on specific tasks. This enables us to make informed decisions about pruning strategies and provides valuable insights into the trade-offs involved in pruning large language models.

\begin{table*}[h]
\centering
\vspace{-0.1in}
\begin{tabular}{ccccccccccc}
\toprule
\multicolumn{2}{c}{\multirow{2}{*}{Task}} &  \multicolumn{9}{c}{Calibration data} \\
& & ARC-C & ARC-E & BoolQ & C4 & PTB & RTE & SST2 & WikiText2 & WNLI \\

\hline
    \multirow{4}{*}{\makecell[c]{Sentiment\\Classification}}
    &Yelp & 70.73 & 72.15 & 73.99 & \textbf{75.03} & 73.03 & 71.83 & 71.30 & 73.50 & 68.77 \\
    &IMDB & 61.99 & 60.99 & 62.91 & \textbf{66.98} & 62.32 & 62.41 & 60.71 & 68.58 & 67.55 \\
    &Sentiment140 & 56.49 & 57.35 & 57.60 & 57.59 & \textbf{57.73} & 56.47 & 53.93 & 57.57 & 56.07 \\
    &SST2 & 66.19 & \textbf{68.80} & 67.22 & 66.95 & 67.07 & 68.53 & 64.54 & 67.88 & 66.60 \\
\midrule
    \multirow{7}{*}{\makecell[c]{Question\\ Answering}}
    &ARC-C & 33.22 & 32.92 & 31.93 & 33.23 & 32.92 & 31.22 & 30.45 & {33.41} & 32.11 \\
    &ARC-E & 65.66 & 65.00 & 64.24 & 65.48 & 64.63 & 62.52 & 62.33 & \textbf{66.22} & 62.83 \\
    &OpenBookQA & 26.17 & 25.95 & 25.75 & \textbf{26.51} & 25.59 & 25.13 & 23.73 & 26.47 & 23.65 \\
    &PubMedQA & 64.00 & 64.05 & 63.31 & 64.28 & \textbf{65.49} & 63.76 & 63.65 & 63.88 & 62.57 \\
    &BoolQ & 70.49 & 69.51 & 70.52 & \textbf{71.81} & 70.76 & 69.49 & 68.94 & 70.47 & 68.34 \\
    &QNLI & \textbf{50.92} & 50.69 & 50.66 & 50.33 & 50.60 & 50.65 & 50.84 & 50.16 & 50.85 \\

\midrule
\multirow{7}{*}{\makecell[c]{Semantic\\ Similarity}}&MRPC & 57.58 & 61.35 & 59.17 & 62.28 & 61.74 & \textbf{62.58} & 57.80 & 61.63 & 60.50 \\
&QQP & \textbf{47.49} & 45.11 & 44.70 & 44.19 & 44.26 & 44.79 & 44.43 & 45.21 & 43.59 \\
&SciTail & 43.77 & 41.59 & 44.74 & 42.66 & 42.84 & \textbf{45.63} & 42.98 & 44.02 & 43.98 \\
&WIC & 50.06 & \textbf{50.30} & 49.90 & 49.78 & 50.15 & 50.24 & 50.02 & 50.12 & 50.11 \\
&MedNLI & \textbf{34.88} & 34.78 & 34.68 & 34.71 & 34.33 & 34.60 & 34.49 & 34.78 & 34.68 \\
&RTE & 58.53 & 57.70 & \textbf{59.01} & 58.52 & 58.74 & 57.94 & 57.87 & 58.70 & 57.56 \\
&PAWS & 46.90 & 46.82 & 46.68 & 46.32 & 46.58 & \textbf{47.46} & 47.07 & 46.95 & 46.03 \\
\midrule
\multirow{6}{*}{\makecell[c]{Logical \\ Reasoning}}&COPA & 79.77 & 80.06 & 79.36 & \textbf{81.13} & 80.00 & 80.15 & 78.30 & 81.01 & 79.20 \\
&LogiQA & 23.32 & \textbf{23.76} & 23.43 & 23.67 & 23.22 & 23.36 & 22.99 & 23.42 & 23.48 \\
&SciQ & 90.63 & 90.57 & 90.53 & 91.17 & \textbf{91.26} & 90.38 & 90.20 & 90.63 & 89.80 \\
&WinoGrande & 63.64 & 63.00 & 63.93 & \textbf{64.98} & 63.95 & 63.37 & 62.59 & 64.33 & 63.71 \\
&WNLI & 47.06 & 46.57 & 46.53 & 46.80 & 47.08 & 46.44 & 46.42 & 46.45 & \textbf{47.27} \\
&WSC273 & 77.31 & 75.86 & 77.96 & 78.63 & \textbf{78.74} & 77.63 & 76.38 & 78.66 & 77.79 \\
&RACE & 37.50 & 37.16 & 37.38 & \textbf{38.41} & 38.34 & 37.35 & 37.23 & 38.21 & 36.59 \\
\bottomrule
\end{tabular}
\vspace{-0.1in}
\caption{Accuracy of different calibration data and tasks averaged over 3 pruning methods and 2 sparsities \{0.25, 05\}.}
\vspace{-0.2in}
\label{all_average}
\end{table*}

\section{Experiments}
\subsection{Calibration Data}
Calibration data is important for pruning methods. 

Table \ref{all_average} displays the accuracy from 24 datasets/tasks (e.g. ARC-E) and 9 calibration datasets (e.g. ARC-C) averaged over 3 pruning methods and 2 sparsities \{0.25, 0.5\}. In addition, Table \ref{tab:Model_cal_Yelp} presents the accuracy of 14 pruned models (including the very new DeepSeek-R1-qwen2-7b \cite{guo2024deepseek}) using the same pruning method Wanda with different calibration data.
\paragraph{Insensitive to Calibration Data} From Table \ref{all_average}, we can see the performance of  Semantic Similarity (e.g., MRPC and QQP) and Logical Reasoning (e.g., SciQ and WNLI) is relatively robust to calibration data. In particular, the performance of SciQ over different calibration sets are very stable and consistent. We believe these tasks likely depend on more general patterns that are less sensitive to variations in calibration data. These findings indicate that pruning methods can maintain performance on these tasks irrespective of the calibration data. 
\paragraph{Sensitive to Calibration Data} From Table \ref{all_average},  Sentiment Classification  like  Yelp and SST2, and Question Answering  like ARC-E, show significant performance variability depending on the calibration data. From Table \ref{tab:Model_cal_Yelp}, in particular, we can clearly see the significant performance variability. For example, the accuracy of the opt-6.7b model on the Yelp dataset reaches 74\% with the WikiText2 calibration data, compared to approximately 60\% with other datasets. Similarly, for the ARC-E task, the DeepSeek-R1-qwen2-7b model shows an accuracy of 66.58\% vs. 35.69\%. with the C4 and SST2 calibration data respectively. 

Calibration data plays a crucial role for the tasks SST2, Yelp, and ARC-E, where different calibration data, such as WikiText2, can lead to a significant improvement in model performance.

In conclusion, this analysis demonstrates that selecting the appropriate calibration data is crucial for optimizing pruning performance. For  sentiment classification and question answering, dataset selection can have a significant impact on model performance. Our findings underscore the importance of understanding the role that calibration data play in pruning LLMs and emphasize the need for further exploration into optimizing  calibration data for different tasks.

\begin{table*}[ht]\small
\centering
\vspace{-0.1in}
\resizebox{2.\columnwidth}{!}{
\begin{tabular}{ccccccccccc}
\toprule
\multirow{2}{*}{Model}& \multirow{2}{*}{Task}&\multicolumn{9}{c}{Calibration Data} \\
 & & ARC-C & ARC-E  & BoolQ & C4 & PTB & RTE & SST2 & WikiText2 & WNLI \\
\midrule
DeepSeek-R1-qwen2-7b &Yelp& 50.04 & 50.04 & 60.53 & \textbf{89.90}  & 82.66 & 50.91 & 50.11 & 87.79 & 50.08 \\
LLaMA-3-8b &Yelp & 75.51 & 76.80 & 50.37 & 95.60 & \textbf{96.90} & 73.64 & 50.00 & 95.85 & 59.79 \\
LLaMA-7b &Yelp& 55.35 & \textbf{63.94} & 57.31 & 50.38 & 50.01 & 61.15 & 53.53 & 50.24 & 57.87 \\
OPT-13b &Yelp& 65.57 & 61.96 & 53.18 & \textbf{70.89} & 55.92 & 53.57 & 63.28 & \textbf{70.89} & 68.04 \\
OPT-30b &Yelp& {83.54} & 74.96 & 73.27 & 71.73 & 71.28 & 79.22 & 74.54 & 61.36 & 86.14 \\
OPT-6.7b &Yelp& 53.65 & 55.38 & 67.43 & 66.37 & 59.47 & 56.54 & 61.17 & \textbf{74.18} & 52.46 \\
Vicuna-7b &Yelp & 88.23 & 85.02 & 80.25 & 95.22 & 92.22 & 91.37 & 53.46 & \textbf{95.47} & 89.14 \\

DeepSeek-R1-qwen2-7b & ARC-E  & 48.86 & 47.47 & 54.84 & \textbf{66.58} & 64.02 & 49.71 & 35.69 & 66.08 & 42.17 \\
LLaMA-3-8b & ARC-E & 62.37 & 60.98 & 62.84 & 70.08 & 69.91 & 59.18 & 40.57 & \textbf{70.66} & 53.79 \\
LLaMA-7b & ARC-E& 62.04 & 61.03 & 61.87 &\textbf{69.70} & 68.18 & 60.86 & 56.02 & 69.65 & 57.49 \\
OPT-13b & ARC-E& 61.28 & 61.15 & 60.77 & \textbf{63.01} & 57.53 & 58.12 & 51.98 & 62.21 & 52.10 \\
OPT-30b & ARC-E& 63.13 & 62.96 & 64.60 & \textbf{68.06} & 64.18 & 61.66 & 51.81 & 66.75 & 52.69 \\
OPT-6.7b & ARC-E& 49.28 & 49.62 & 50.88 & \textbf{57.45} & 53.75 & 49.45 & 50.00 & 56.40 & 45.66 \\
Vicuna-7b & ARC-E & 64.65 & 65.57 & 66.20 & 69.23 & \textbf{70.16} & 64.69 & 60.98 & 69.23 & 62.37 \\
\bottomrule

\end{tabular}
}
\vspace{-0.05in}
\caption{Accuracy of pruned models across different calibration data and evaluation on Yelp and ARC-E dataset using Wanda pruning.}
\vspace{-0.2in}
\label{tab:Model_cal_Yelp}
\end{table*}

\subsection{Neuron Semantic Attribution (NSA)}
\label{sec:expNSA}
\paragraph{Motivation}
For sentiment classification like Yelp in Table \ref{tab:Model_cal_Yelp}, we observe a noticeable performance variance in certain models after pruning. This variance in performance prompted us to explore the underlying reasons behind it. The core of this exploration is to understand what has been pruned—which components of the model are being removed, and how this removal affects its ability to perform sentiment classification. To investigate this, we applied our analysis method, Neuron Semantic Attribution (NSA), to analyze the relationship between tokens and the corresponding neuron activations. Specifically, we sampled texts from the Yelp dataset. We visualized how the tokens are connected to the neurons in the model. The method we used allows us to examine which parts of the text influence the model’s decision-making process.
\paragraph{NSA Visualization on Yelp Data}
In Figure \ref{fig:Yelp_neuron}, we show an analysis example of this NSA method. The depth of the color represents the activation strength of the neurons connected to the tokens. We observe that the color intensity of the highlighted words changes significantly between the unpruned and pruned models. In addition, we can see that the activation of key sentiment-related tokens, such as `badly' showed a decrease in activation strength after pruning. This suggests that the pruning  weakens of the neurons responsible for sentiment analysis,  explaining the performance decline.

Interestingly, although there is a decrease in activation for the pruned model, the most influential tokens still retain a significant level of activation, which explains why the pruned models can still maintain reasonably good performance for sentiment classification. However, the reduced activation strength may indicate that the model is less confident in its decision-making process, potentially contributing to the observed performance gap.


\subsection{Component  Analysis}
\label{sec:Component_Analysis}
\begin{figure}[ht] 
\vspace{-0.25in}
    \centering
    \includegraphics[width=1.0\linewidth]{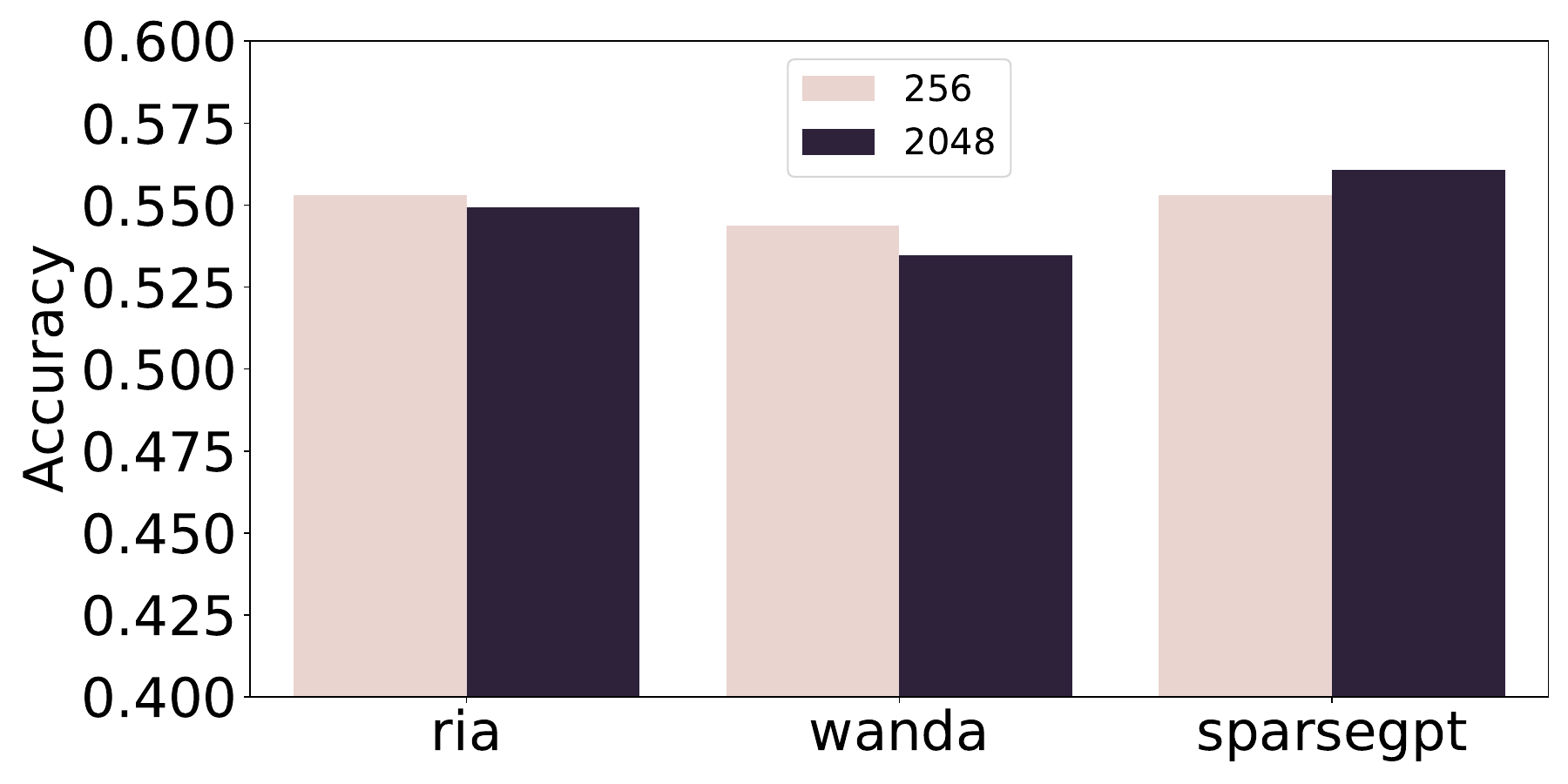}
    \vspace{-0.3in}
    \caption{The average accuracy of pruned model with different sequence length of tokens.}
    \label{fig:bar_Method_Seq}
    \vspace{-0.3in}
\end{figure}

\paragraph{Sequence Length of Tokens} To further investigate the impact of sequence length of tokens on the performance of pruned models, we plotted a bar chart in Figure \ref{fig:bar_Method_Seq} comparing the three pruning methods. The results suggest that, for the pruned models in this study,
sequence length has little impact on performance. More detailed experiments on sequence length can be found in our supplementary materials. 

\begin{figure}[ht] 
\vspace{-0.1in}
    \centering
    \includegraphics[width=1.0\linewidth]{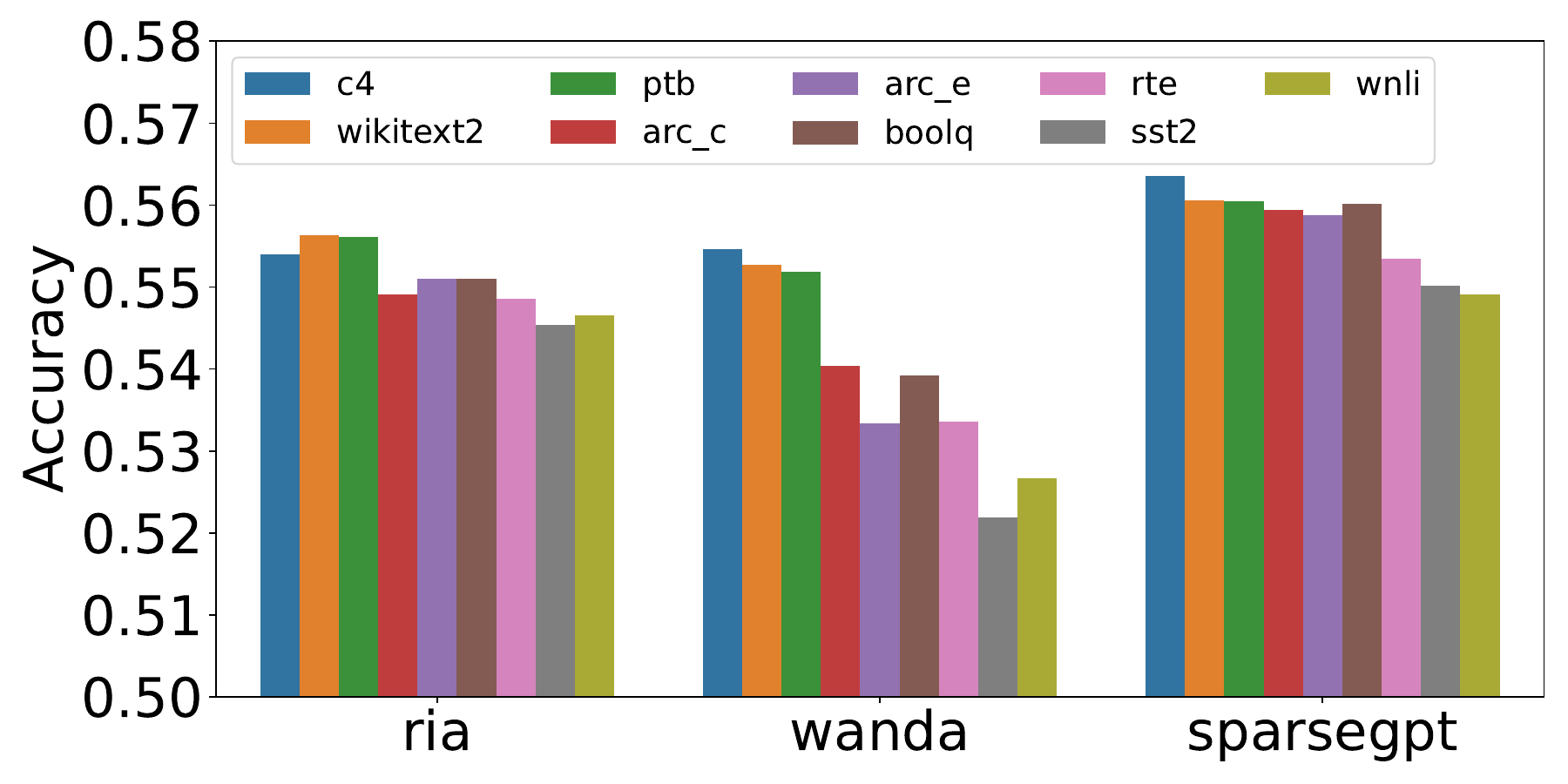}
    \vspace{-0.4in}
    \caption{The  accuracy of three pruning methods using 9 different calibration data averaged over 24 tasks.}
    \vspace{-0.35in}
    \label{fig:bar_Method_Cal}
\end{figure}

\paragraph{Pruning Methods} In Figure \ref{fig:bar_Method_Cal}, we compare the accuracy of the three pruning methods using different calibration data averaged over 24 tasks. SparseGPT demonstrates the most competitive performance on most calibration data. 

These extensive evaluations highlight the importance of empirical evaluation in identifying the most effective method for specific tasks.

\begin{figure}[ht] 
    \centering
    \includegraphics[width=1.0\linewidth]{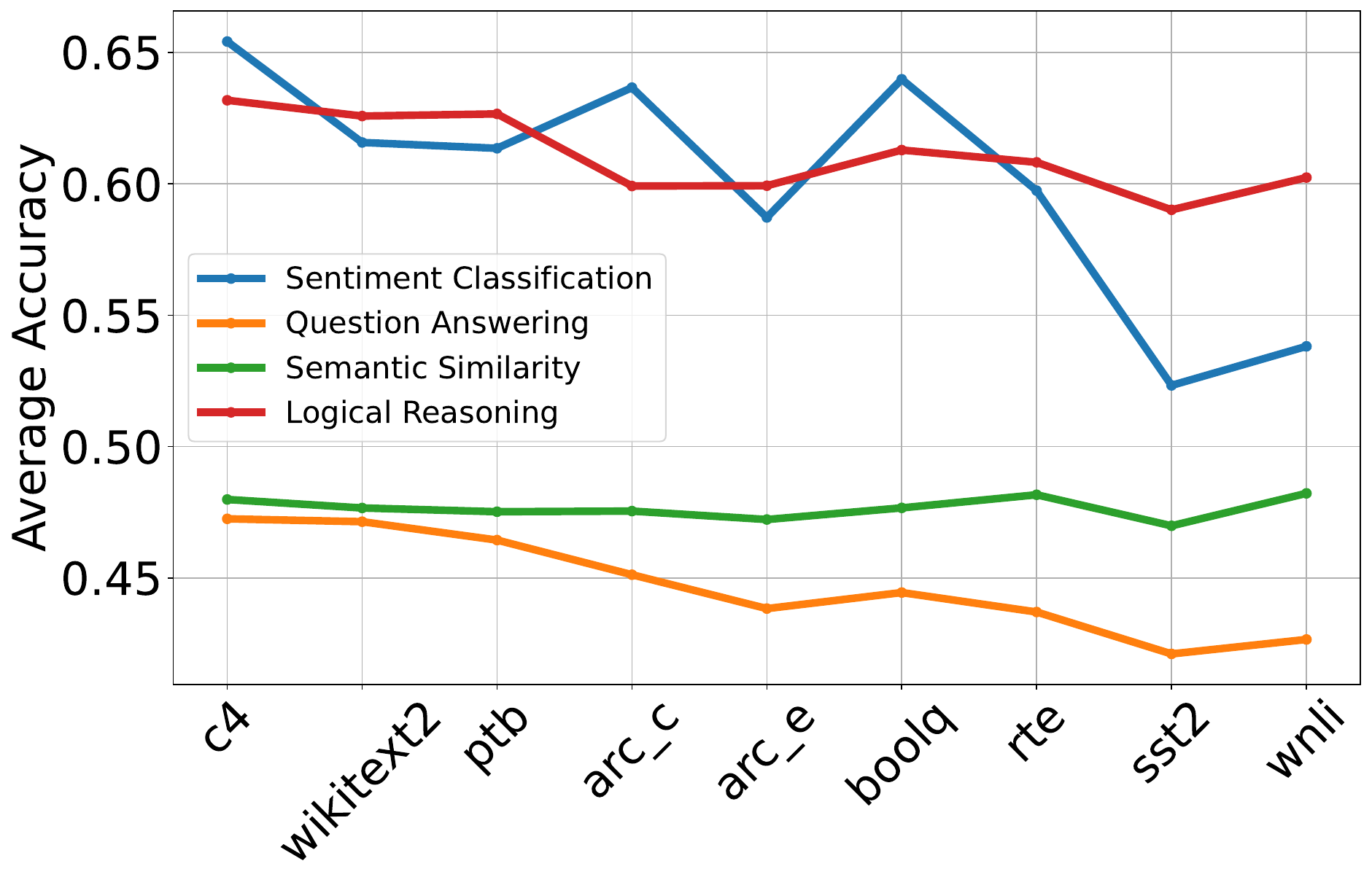}
    \vspace{-0.3in}
    \caption{Accuracy of pruned models on different calibration data with 2:4 sparsity averaged over 4 kinds of tasks.}
    \label{fig:cal_tasks}
    \vspace{-0.2in}
\end{figure}

\paragraph{Downstream Tasks} We also conducted a thorough analysis of the performance differences across four downstream tasks. In Figure \ref{fig:cal_tasks}, our  results show that Sentiment Classification tasks exhibit significant performance variations across different calibration data, demonstrating a high degree of sensitivity.  The performance variations are explained by our NSA visualization in Section \ref{sec:expNSA}.

In contrast, Question Answering and Logical Reasoning tasks show relatively stable performance. Specifically, for tasks such as those in the ARC-C and ARC-E datasets, the model’s performance does not exhibit significant variations with changes in calibration data. This suggests that these types of tasks rely more on reasoning processes. These tasks generally involve extracting key information from provided contexts, and this reasoning ability depends less on the precise composition of the calibration data, making them more robust to such changes.
\paragraph{Sparsity Ratio} The results in Table \ref{tab:task_sparse} highlight the impact of varying sparsity ratios on the performance of pruned models across multiple tasks. In general, lower sparsity ratios (e.g., 0.1 or 0.2) retain a higher proportion of model parameters, leading to better performance on most tasks. ARC-E achieves a high accuracy of 76.43 at a sparsity ratio of 0.1, which gradually declines to 27.30 at a sparsity ratio of 0.8. BoolQ similarly shows a consistent drop from 77.53 (sparsity ratio 0.1) to 37.86 (sparsity ratio 0.8).

Interestingly, certain tasks such as IMDB and PAWS exhibit more stable performance across different sparsity ratios, indicating their robustness to parameter pruning. For instance, PAWS maintains accuracy around 44 across sparsity ratios, with slight improvements at higher sparsity levels (e.g., 55.75 at sparsity ratio 0.8). On the other hand, tasks like SciQ show a sharp performance drop as sparsity increases. The accuracy remains high (93.80 to 94.45) for sparsity ratios between 0.1 and 0.4, but drastically decreases to 28.00 at 0.8.

Surprisingly, we find that some scenarios have higher accuracy with a higher sparsity ratio, e.g.  SciTail. We carefully checked our implementation and we do not think it is caused by wrong implementation. To our knowledge, this finding has not been investigated by the society. We hope that this finding can inspire researchers explore this direction deeply.

\begin{table}[ht]
\resizebox{1.\columnwidth}{!}{
\begin{tabular}{ccccccccc}
\toprule
\multirow{2}{*}{Task} &\multicolumn{8}{c}{Sparsity ratio} \\
& 0.1 & 0.2 & 0.3 & 0.4 & 0.5 & 0.6 & 0.7 & 0.8 \\
\midrule
ARC-C & 43.25 & 43.17 & 42.62 & 41.17 & 36.82 & 28.41 & 18.47 & 20.39 \\
ARC-E & 76.43 & 76.31 & 75.76 & 73.91 & 69.88 & 60.71 & 30.89 & 27.30 \\
BoolQ & 77.53 & 77.75 & 76.88 & 74.78 & 72.59 & 65.96 & 49.57 & 37.86 \\
COPA & 87.00 & 88.50 & 86.50 & 85.00 & 84.00 & 75.00 & 65.00 & 62.00 \\
IMDB & 50.30 & 50.30 & 50.34 & 50.36 & 50.98 & 55.97 & 51.73 & 51.02 \\
LogiQA & 25.88 & 24.73 & 22.66 & 22.04 & 23.88 & 22.12 & 20.05 & 19.66 \\
MedNLI & 34.41 & 34.41 & 33.37 & 33.62 & 34.38 & 33.30 & 34.98 & 33.37 \\
MRPC & 69.24 & 68.75 & 65.69 & 62.26 & 54.41 & 55.89 & 49.75 & 31.62 \\
OpenBookQA & 31.30 & 32.50 & 31.70 & 30.60 & 28.30 & 22.90 & 12.40 & 14.20 \\
PAWS & 44.24 & 44.24 & 44.25 & 44.24 & 44.24 & 44.30 & 50.33 & 55.75 \\
PubMedQA & 71.50 & 71.20 & 70.30 & 70.60 & 70.00 & 58.80 & 41.40 & 33.80 \\
QNLI & 49.89 & 50.33 & 50.36 & 49.98 & 50.09 & 49.58 & 50.98 & 50.48 \\
QQP & 37.13 & 38.37 & 39.14 & 37.02 & 37.21 & 37.48 & 55.67 & 63.12 \\
RACE & 39.66 & 39.47 & 39.95 & 40.19 & 39.19 & 34.50 & 25.45 & 23.01 \\
RTE & 61.73 & 63.18 & 63.36 & 58.84 & 54.69 & 52.89 & 52.71 & 51.98 \\
SciQ & 93.80 & 94.00 & 94.30 & 94.45 & 94.15 & 91.00 & 76.40 & 28.00 \\
SciTail & 45.59 & 45.59 & 42.52 & 35.43 & 33.55 & 49.31 & 49.62 & 49.62 \\
Sentiment140 & 50.32 & 50.32 & 50.15 & 50.30 & 56.46 & 57.15 & 51.45 & 50.89 \\
SST2 & 49.48 & 49.25 & 49.14 & 49.14 & 49.54 & 59.75 & 49.83 & 51.38 \\
WIC & 49.92 & 49.84 & 49.14 & 49.22 & 48.20 & 49.53 & 49.38 & 50.00 \\
WinoGrande & 69.22 & 69.53 & 68.90 & 68.78 & 66.85 & 61.60 & 50.63 & 49.73 \\
WNLI & 44.36 & 46.48 & 41.55 & 45.77 & 42.25 & 43.66 & 43.66 & 49.30 \\
WSC273 & 80.59 & 80.95 & 82.23 & 82.78 & 80.59 & 73.44 & 55.13 & 50.91 \\
Yelp & 50.02 & 50.03 & 50.02 & 50.05 & 52.44 & 54.23 & 52.58 & 52.76 \\
\bottomrule
\end{tabular}}
\vspace{-0.1in}
\caption{Accuracy across  tasks at different sparsity ratios. We use llama-7b model, RIA pruning, and C4  calibration data.}
\vspace{-0.1in}
\label{tab:task_sparse}
\end{table}

\begin{table}[ht]
\centering
\resizebox{1.\columnwidth}{!}{
\begin{tabular}{cccccccccc}
\toprule
\multirow{2}{*}{Model} &\multicolumn{9}{c}{Calibration Data} \\
 & ARC-C & ARC-E & BoolQ & C4 & PTB & RTE & SST2 & WikiText2 & WNLI \\
\midrule
DeepSeek-R1-qwen2-7b & 49.12 & 49.12 & 49.22 & \textbf{53.10} & \textbf{53.10} & 48.81 & 48.61 & 52.50 & 48.45 \\
LLaMA-13b & 59.55 & 60.42 & 60.78 & 60.55 & 60.49 & 61.25 & 58.75 & \textbf{61.27}& 58.88 \\
LLaMA-3-8b & 55.91 & 55.86 & 55.95 & \textbf{57.29} & 57.19 & 55.65 & 54.40 & 57.19 & 55.33 \\
LLaMA-7b & 52.54 & 52.01 & 52.09 & 52.55 & \textbf{53.26} & 51.86 & 51.99 & 52.46 & 51.99 \\
Mistral-7b & 53.45 & 53.46 & 53.37 & \textbf{54.62} & 54.42 & 52.86 & 52.52 & 54.33 & 52.77 \\
OPT-13b & 54.11 & 53.53 & 53.92 & \textbf{54.35} & 53.47 & 53.78 & 54.27 & 53.33 & 53.80 \\
OPT-30b & 53.91 & 53.58 & 53.95 & \textbf{54.73} & 54.03 & 53.81 & 52.53 & 54.41 & 53.33 \\
OPT-6.7b & 52.82 & 52.96 & 53.28 & \textbf{54.05} & 53.14 & 52.31 & 52.27 & 53.64 & 52.13 \\
Vicuna-13b & 59.00 & 59.00 & 59.56 & \textbf{61.06} & 60.62 & 58.73 & 58.41 & 60.79 & 57.98 \\
Vicuna-7b & 58.34 & 58.27 & 58.17 & 58.78 & \textbf{59.00} & 57.35 & 56.60 & 58.69 & 57.13 \\
\bottomrule
\end{tabular}
}
\vspace{-0.1in}
\caption{Accuracy of pruned models on different calibration data with 2:4 sparsity averaged over 3 pruning methods and 24.}
\vspace{-0.2 in}
\label{tab:cal_model}
\end{table}

\paragraph{Cross-Model Pruning} It is also interesting to know how  each LLM performs over different calibration data. 
Table \ref{tab:cal_model} shows the impact of different calibration data on pruning results for each model. For most models, datasets like C4 and WikiText2 consistently lead to higher performance after pruning, suggesting they are more effective at preserving model capacity. For instance, llama-13b performs best on WikiText2 (61.27) and RTE (61.25), while vicuna-13b achieves its highest score on C4 (61.06). Conversely, datasets such as WNLI and ARC-C often result in lower performance, indicating their limited effectiveness for pruning calibration data.

Overall, the impact of calibration data varies across models. Larger models like LLaMA-13b and Vicuna-13b exhibit relatively stable performance across datasets, while smaller models like Mistral-7b and OPT-6.7b show greater variability, reflecting a stronger dependence on the choice of calibration data. This suggests that selecting appropriate datasets, such as C4 or WikiText2, can improve pruning outcomes, especially for smaller models.

\section{Conclusion}
In this paper, we explore the generalizability of  pruning methods on large language models using three methods—SparseGPT, Wanda, and RIA—across various tasks. Our findings show that pruning methods reduce model size without significant performance loss, but the choice of calibration dataset, pruning sparsity, and task type influence outcomes. Sentiment classification is more sensitive to pruning methods and calibration datasets, while tasks like question answering and logical reasoning are more robust. Using Neuron Semantic Attribution (NSA), we demonstrate how pruning methods impact neuron activations and model decision-making. Our work provides insights into the task-specific nature of pruning and lays the foundation for future research to develop more efficient and interpretable pruning strategies.

\nocite{langley00}

\bibliography{main}
\bibliographystyle{icml2025}


\newpage
\appendix
\onecolumn

\section{Appendix}

\begin{figure*}[ht] 
    \centering
    \includegraphics[width=1.0\linewidth]{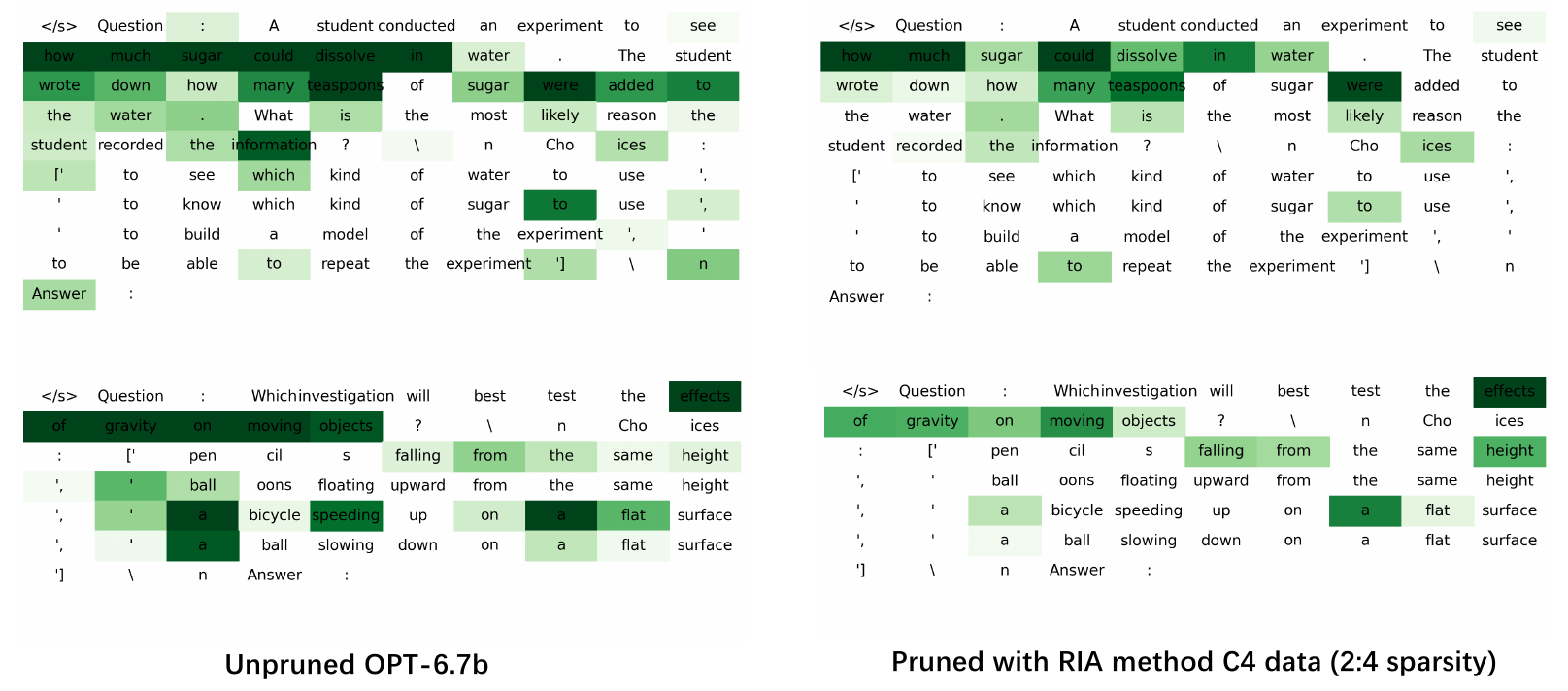}
    \vspace{-0.3in}
    \caption{The activation of neuron 8217 and neuron 8283 in layer 23.The pruned model is OPT-6.7b with RIA method and calibration dataset C4. Deeper
colors in the visualization represent stronger neuron activations for specific tokens.}
    \label{fig:arc_challenge_neuron}
\end{figure*}

\begin{figure*}[ht]\small 
    \centering
    \includegraphics[width=1.0\linewidth]{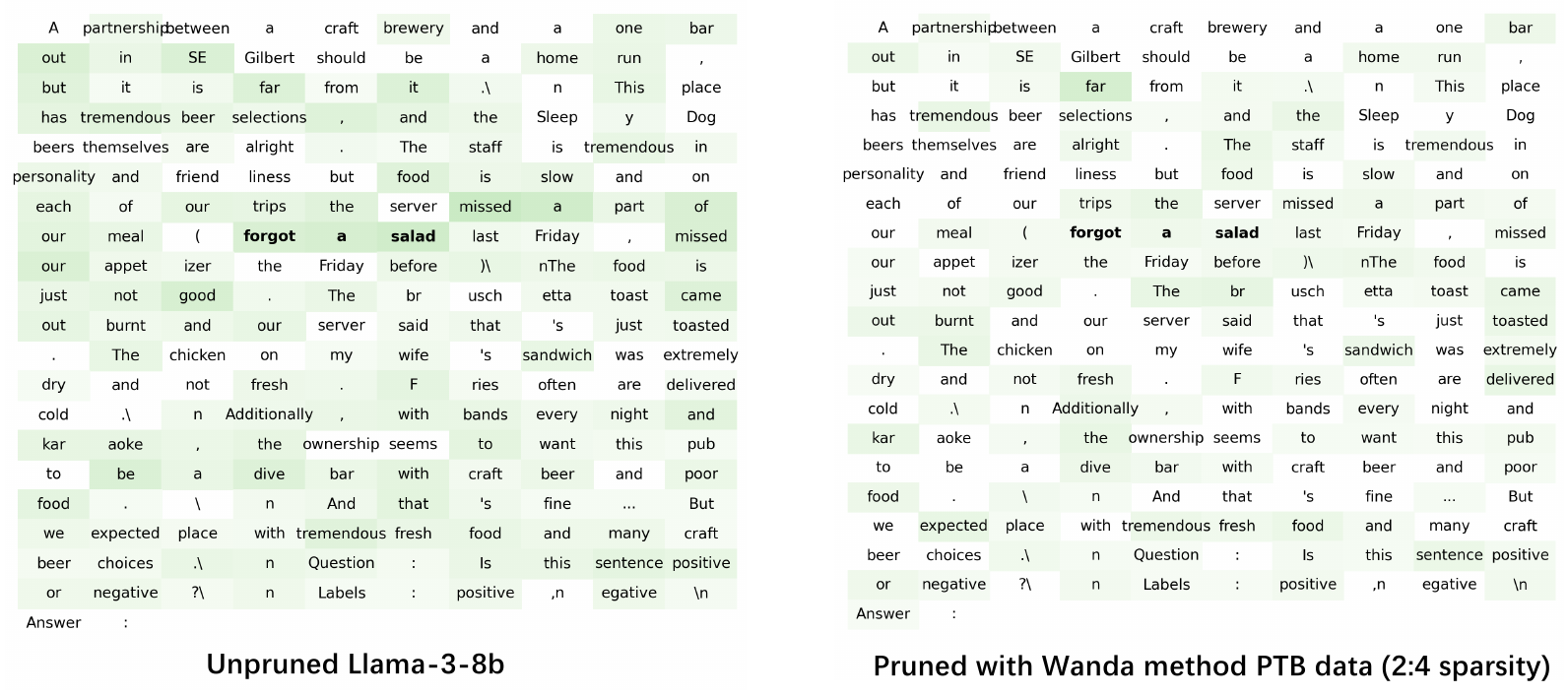}
    \vspace{-0.3in}
    \caption{The activation of neuron 2287 in layer 23.The pruned model is llama-3-8b with Wanda method and calibration dataset PTB. Deeper
colors in the visualization represent stronger neuron activations for specific tokens.}
    \label{fig:imdb_neuron}
\end{figure*}

\textbf{NSA Visualization on ARC-C Data}

We also applied NSA to the ARC-C dataset, as shown in Fig.\ref{fig:arc_challenge_neuron}, which presented a different result.

In the case of the ARC-C dataset, the results were quite different. The visualization in Fig.\ref{fig:arc_challenge_neuron} reveals that, after pruning, the model retains a strong focus on key tokens related to the task, such as words representing question cues or common reasoning patterns. Unlike sentiment classification, where pruning caused a noticeable reduction in activation strength, the impact on the ARC-C task was less pronounced. The tokens connected to reasoning and question-answering were still highly activated post-pruning, suggesting that the model maintained its ability to reason through the problem.

In fact, the intensity of activation in ARC-C remained relatively stable across both the original and pruned models. This suggests that, unlike sentiment classification, tasks like ARC-C may be less sensitive to pruning, possibly due to the inherent structure of reasoning tasks. In this dataset, pruning appears to have less impact on the model’s ability to focus on critical decision-making tokens, which might explain why the performance drop was less significant compared to tasks like sentiment classification.

\textbf{NSA Visualization on IMDB Data}

Figure \ref{fig:imdb_neuron} presents the activation of neuron 2287 in layer 23 of the pruned model (Llama-3-8b) using the Wanda method and the PTB calibration dataset.

\textbf{Other Analysis}

Table \ref{tab:task_seq} shows that performance varies across different sequence lengths. For tasks like ARC-E and ARC-C, shorter sequences (e.g., 64) yield higher accuracy, while longer sequences (e.g., 2048) tend to decrease performance. Conversely, tasks like MRPC and SciQ benefit from longer sequences, with MRPC’s accuracy increasing from 44.12 at seqlen 64 to 72.79 at seqlen 2048. Sentiment tasks like Yelp and SST2 perform best with shorter sequences, suggesting that excessively long sequences might not always be beneficial for certain tasks.
\begin{table*}[ht]
\centering
\resizebox{1.\textwidth}{!}{
\begin{tabular}{ccccccccccccc}
\toprule
Seqlen & ARC-C & ARC-E & BoolQ & COPA & IMDB & LogiQA & MedNLI & MRPC & OpenBookQA & PAWS & PubMedQA & QNLI \\
\midrule
64 & \textbf{38.57} & \textbf{71.17} & 75.23 & 82.00 & \textbf{76.75} & 24.12 & 33.19 & 44.12 & 28.00 & 45.77 & 69.60 & 50.85 \\
128 & 38.23 & 70.96 & 75.35 & \textbf{83.00} & 62.70 & 23.96 & 33.33 & 60.05 & 28.60 & 44.36 & \textbf{70.20} & 51.33 \\
256 & 37.63 & 71.13 & \textbf{75.90} & 77.00 & 63.88 & 24.73 & 33.33 & 54.90 & \textbf{29.00} & 44.72 & 69.40 & \textbf{51.84} \\
512 & 37.12 & 70.41 & 74.40 & 80.00 & 64.70 & 24.73 & 33.33 & 46.08 & 27.00 & \textbf{46.17} & 68.60 & 50.92 \\
1024 & 35.92 & 69.70 & 73.30 & 72.00 & 68.93 & \textbf{24.88} & \textbf{33.41} & \textbf{72.79} & 25.20 & 44.72 & 67.00 & 50.69 \\
2048 & 32.00 & 63.38 & 57.83 & 73.00 & 66.77 & 21.97 & 33.33 & 35.54 & 21.60 & 44.24 & 64.80 & 50.41 \\
\midrule
Seqlen & QQP & RACE & RTE & SciQ & SciTail & Sentiment140 & SST2 & WIC & WinoGrande & WNLI & WSC273 & Yelp \\
\midrule
64 & 65.25 & 40.38 & \textbf{66.06} & 92.90 & 48.85 & \textbf{71.46} & \textbf{79.93} & 50.16 & \textbf{66.14} & 43.66 & 76.92 & \textbf{75.94} \\
128 & \textbf{65.93} & \textbf{40.48} & 65.34 & \textbf{93.00} & 49.23 & 65.62 & 75.92 & 50.16 & 65.27 & 43.66 & 78.02 & 64.97 \\
256 & 65.90 & 40.00 & 62.09 & 92.90 & \textbf{49.54} & 63.80 & 71.44 & \textbf{50.47} & 65.35 & \textbf{46.48} & 77.29 & 66.94 \\
512 & 63.51 & 39.33 & 61.73 & 92.90 & 49.54 & 65.95 & 76.83 & 50.16 & 63.54 & 45.07 & \textbf{80.22} & 70.04 \\
1024 & 63.83 & 38.66 & 62.09 & 92.90 & 48.16 & 68.44 & 76.95 & 50.00 & 62.27 & 42.25 & 78.02 & 74.70 \\
2048 & 62.98 & 36.65 & 58.48 & 91.10 & 42.56 & 61.18 & 63.19 & 50.00 & 56.67 & 45.07 & 72.89 & 68.08 \\
\bottomrule
\end{tabular}}
\caption{Performance across various tasks at different sequence lengths.}
\vspace{-0.2in}
\label{tab:task_seq}
\end{table*}


\end{document}